\documentclass{article}

\usepackage{microtype}
\usepackage{graphicx}
\usepackage{subfigure}
\usepackage{booktabs} 

\usepackage{hyperref}

\usepackage[accepted]{icml2025}

\usepackage{amsmath}
\usepackage{amssymb}
\usepackage{mathtools}
\usepackage{amsthm}
\usepackage{array}

\usepackage{amsmath}
\usepackage{amssymb}
\usepackage{amsfonts}
\usepackage{titlesec}
\usepackage{graphicx}
\usepackage{paralist}
\usepackage{listings}
\usepackage{enumitem}
\usepackage{authblk}
\usepackage{float}
\usepackage{subcaption}

\usepackage{booktabs}
\usepackage{xcolor}
\usepackage{colortbl}
\usepackage{array}

\definecolor{light_blue}{rgb}{0.7, 0.885, 0.988}
\definecolor{lighter_blue}{rgb}{0.8, 0.92, 0.99}
\definecolor{lightest_blue}{rgb}{0.9, 0.96, 0.995}

\usepackage{amsmath}
\usepackage{amssymb}
\usepackage{amsthm}
\usepackage{mathtools}
\theoremstyle{plain}

\theoremstyle{definition}

\icmltitlerunning{A Single Direction of Truth}

\begin{document}

\twocolumn[
\icmltitle{A Single Direction of Truth: An Observer Model’s Linear Residual Probe Exposes and Steers Contextual Hallucinations}




\begin{icmlauthorlist}
\icmlauthor{Charles O'Neill}{yyy}
\icmlauthor{Slava Chalnev}{comp}
\icmlauthor{Chi Chi Zhao}{yyy}
\icmlauthor{Max Kirkby}{yyy}
\icmlauthor{Mudith Jayasekara}{yyy}
\end{icmlauthorlist}

\icmlaffiliation{yyy}{Parsed, London, UK}
\icmlaffiliation{comp}{Independent}

\icmlcorrespondingauthor{Charles O'Neill}{charles@parsed.com}

\icmlkeywords{mechanistic interpretability, language models, hallucinations}

\vskip 0.3in
]



\printAffiliationsAndNotice{}  

\begin{abstract}
Contextual hallucinations –- statements unsupported by given context –- remain a significant challenge in AI. We demonstrate a practical interpretability insight: a generator-agnostic observer model detects hallucinations via a single forward pass and a linear probe on its residual stream. This probe isolates a single, transferable linear direction separating hallucinated from faithful text, outperforming baselines by 5–27 points and showing robust mid‑layer performance across Gemma‑2 models (2B$\to$27B). Gradient‑times‑activation localises this signal to sparse, late‑layer MLP activity. Critically, manipulating this direction causally steers generator hallucination rates, proving its actionability. Our results offer novel evidence of internal, low‑dimensional hallucination tracking linked to specific MLP sub‑circuits, exploitable for detection and mitigation. We release the 2000‑example \textsc{ContraTales} benchmark for realistic assessment of such solutions.
\end{abstract}

\section{Introduction}
\label{sec:introduction}

Large language models (LLMs) have made striking progress in open‑ended generation, yet they continue to produce statements that are \emph{plausible but unsupported} by their given context -- a failure mode broadly labelled \textit{hallucination} \citep{Huang2024Survey, ji2023survey}. When these systems are deployed in medicine, finance, or law, even an isolated hallucination can trigger costly or harmful downstream actions, eroding user trust and slowing adoption. Detecting hallucinations \emph{after} text has been produced is therefore a key practical requirement, both for automated self‑monitoring and for post‑hoc auditing workflows \citep{Valentin2024CostEffectiveHallucinationDetection}.

Unfortunately, most high‑performing detectors to date either (i) require privileged access to the \textbf{generator model}’s parameters and logits -- an unrealistic assumption for commercial APIs and safety‑critical settings -- or (ii) fall back on brittle surface cues such as lexical overlap, unseen named entities, or low embedding similarity between source and output. These heuristics falter whenever hallucinations manifest as subtle logical contradictions rather than obvious factual novelties, a gap that is starkly revealed by synthetic logic benchmarks such as our \textsc{ContraTales}. This highlights the need for robust, interpretability-based approaches that offer deeper insights beyond surface cues for more reliable detection

Recent work in mechanistic interpretability argues that transformers internally encode rich world knowledge along \emph{approximately linear} directions in activation space \citep{Burns2023DiscoveringLatentKnowledge, park2023linear}. Linear or sparse probes applied to hidden states have uncovered directions for factual truth, policy compliance, and other high‑level properties \citep{alain2016understanding, marks2024sparse}. Yet, existing studies either rely on signals \emph{inside} the generator or do not test whether the discovered directions are (i) \emph{consistent} across model sizes and domains, (ii) \emph{causal}—that is, capable of steering generation when manipulated—and (iii) sufficiently salient to be decoded by small observer models, limiting their direct actionability in practical, generator-agnostic scenarios.

We ask: does a transformer \emph{notice} when a span of text contradicts its earlier context, and if so, is that realisation encoded along a single, linearly accessible axis that can be read out by another model? We cast this as a mechanistic hypothesis and test it directly. Concretely, we introduce the \textbf{observer paradigm}: a \emph{separate}, \emph{frozen} language model ingests an arbitrary document consisting of a source passage followed by a candidate continuation and, in a single forward pass, predicts whether the continuation is contextually supported. Our detector uses nothing more than a linear probe on the observer’s residual‑stream activations at the full stop of the final sentence.

Across standard news‑domain summarisation sets (\textsc{CNN/DailyMail}, \textsc{XSum}) and an out‑of‑domain corpora (our introduced dataset \textsc{ContraTales}) our linear probe approach achieves F\textsubscript{1} scores up to 0.99 on news and 0.84 on \textsc{ContraTales}, significantly beating lexical‑overlap, entity‑verification, semantic‑similarity, and attention‑pattern baselines by 5–27 points. Layer sweeps reveal a broad mid‑layer performance plateau that is \emph{shared} across model sizes (Gemma-2 2B$\to$27B) and datasets. Attribution analysis based on gradient‑times‑activation localises the signal for contextual inconsistency to a sparse pattern of late‑layer MLP activity, whose characteristics are notably stable across diverse datasets. Most importantly, causal interventions that ablate or inject the learned direction in a generator modulate hallucination rates, demonstrating that the axis captures a functionally meaningful representation.

\paragraph{Contributions.}
\begin{enumerate}
\item We provide the first \emph{generator‑agnostic} detector that identifies contextual hallucinations with a \emph{single forward pass}, requiring no knowledge of the text’s origin and no sampling overhead, demonstrating a practical application of interpretability for efficient AI monitoring.
\item We show that a \emph{single, highly linear} residual‑stream direction robustly separates hallucinated from supported spans across domains and remains readable by comparatively small observer models, offering an actionable insight into model representations.
\item Through gradient‑times‑activation attribution, we map the representation of contextual inconsistency to a compact and consistent pattern of late‑layer MLP activity, and demonstrate its sparsity and stability across key datasets.
\item We establish \emph{causality}: manipulating activity along this axis in a generator steerably reduces or amplifies hallucination prevalence.
\item To contribute to developing realistic benchmarking methods for actionable interpretability, we release \textsc{ContraTales}, a 2000‑example benchmark of purely logical contradictions tailored to stress‑test contextual hallucination detectors.
\end{enumerate}

\section{Background and Related Work}

Large language models (LLMs) frequently generate plausible yet factually incorrect content that contradicts their input or world knowledge, a phenomenon known as hallucination \citep{Huang2024Survey, ji2023survey}. Despite remarkable capabilities in text generation and comprehension, this tendency to hallucinate significantly undermines model reliability in high-stakes domains \citep{ji2023survey}. We focus on \textit{intrinsic hallucinations}, which directly contradict provided source content, as opposed to \textit{extrinsic hallucinations}, which introduce unverifiable external information \citep{ji2023survey, Huang2024Survey}.\footnote{Readers new to mechanistic interpretability (MI) will find a concise overview in \citet{Rai2025SurveyMI}.}

\subsection{Hallucination Detection Methodologies}

Existing detection methods can be categorised by their access requirements to the generating model and by underlying techniques.

Generator-internal approaches require access to the generator’s internals, encompassing token-probability analysis, calibrated confidence estimation, and internal-state probing \citep{Azaria2023InternalState}. Recent MI-based work extends this line, with \citet{Yu2024MechanisticHalluc} identifying drifting sub-modules for non-factual outputs, and \citet{Sun2025ReDeEP} tracing retrieval versus parametric knowledge to flag hallucinations in RAG systems. However, such methods are inapplicable to black-box APIs \citep{Valentin2024CostEffectiveHallucinationDetection}. Post-hoc external methods operate on generated text alone, and include checking claims against knowledge bases, using entailment models, or enlisting an LLM judge \citep{Huang2024Survey, Valentin2024CostEffectiveHallucinationDetection}. Sampling-based consistency, such as SelfCheckGPT \citep{Manakul2023SelfCheckGPT}, generates and compares multiple outputs for agreement but is computationally intensive.

Embedding/representation-based methods exploit vector representations or internal states without needing generator weight access. Examples include semantic similarity checks of source and output embeddings \citep{ji2023survey}, internal-state distribution analysis for hidden-state drift \citep{Farquhar2024SemanticEntropy}, and efficient linear classifiers like Semantic Entropy Probes (SEPs) \citep{Kossen2024SemanticEntropyProbes}. Hallucination-resistant finetuning along ``hallucination directions'' has also been demonstrated \citep{HalluShield2025}. The ongoing challenge is to design detectors that are simultaneously efficient, post-hoc, and generator-agnostic.

\subsection{Utilising LLM Internals}

\paragraph{Linear Representation Hypothesis}
The \emph{linear representation hypothesis} (LRH) posits that a transformer’s activation space can be approximated as a sparse sum of \emph{linearly separable feature directions}, first articulated for toy models by \citet{Elhage2022Toy} and formalised for language models by \citet{park2023linear}. Empirical support includes sparse-autoencoder (SAE) studies showing that few orthogonal, near-monosemantic directions can reconstruct most hidden-state variance \citep{makhzani2013k,cunningham2023sparse}, and linear or low-rank probes isolating causal directions for truthfulness, gender bias, and chain-of-thought features \citep{alain2016understanding,nanda2023emergent,marks2024sparse}. Nonetheless, competing evidence, such as observations of multidimensional toroidal embeddings in Llama and Mistral \citep{engels2024not}, suggests the LRH may be incomplete.

\paragraph{Probing}
Probing operationalises the LRH by training lightweight decoders on hidden states to predict an external label. Originating in visual-cortex studies using linear classifiers on biological neurons \citep{mur2009revealing}, it was adapted to ANNs by \citet{alain2016understanding} and is now an interpretability research staple. A growing body of work demonstrates that concepts like factual truth \citep{Burns2023DiscoveringLatentKnowledge}, policy compliance \citep{Azaria2023InternalState}, and even sleeper-agent triggers \citep{hubinger2024sleeper} are linearly decodable, especially from middle-to-late layers, consistent with transformer-circuits analyses \citep{Olah2020TransformerCircuits}. In the context of hallucination, \citet{Azaria2023InternalState} also investigated internal states for related properties. \citet{simhi2024constructing} demonstrated probe use across most layers and components in a 7B model, though with limited transferability and a narrow focus on unambiguous answers. While many approaches link hallucination to model uncertainty \citep{Farquhar2024SemanticEntropy}, models can also exhibit high-certainty hallucinations \citep{Simhi2025TrustMeImWrong}. Other relevant research has shown LLMs internally encode question-answerability \citep{slobodkin2023curious}, or has emphasised latent-knowledge awareness \citep{ferrando2024know} and controllable context reliance \citep{minder2024controllable}.

\paragraph{Attention} Attention patterns, which reveal information flow, can also detect contextual hallucinations. For example, Lookback Lens quantifies the balance between context-focused and self-focused heads to detect contextual hallucinations \citep{Chuang2024LookbackLens}. Similar attention-based mechanisms underpin causal editing efforts by \citet{ferrando2024know}, controllable context-sensitivity work by \citet{minder2024controllable}, and \citet{yuksekgonul2024attentionsatisfiesconstraintsatisfactionlens} found a strong positive correlation between an LLM's attention to relevant prior tokens and the factual accuracy of its generations.

\section{Methods}
\label{sec:methods}

This section details the datasets, our proposed linear probing methodology for detecting contextual hallucinations, the baseline methods used for comparison, and the unit-level attribution technique employed to interpret the probe.

\subsection{Datasets}
\label{sec:datasets}

Our core task is detecting contextual inconsistencies. In our observer paradigm, this involves a model identifying text spans within a concatenated input-output sequence that either contradict or lack support from the preceding context. The observer model processes this unified document to detect such unsupported claims. We evaluate this capability using four datasets, summarised in Table~\ref{tab:datasets}.

\begin{table*}[t]
\scriptsize
\centering
\caption{Summary of datasets used for evaluating hallucination detection.}
\label{tab:datasets}
\begin{tabular}{@{}lllllr@{}}
\toprule
\textbf{Dataset} & \textbf{Input} & \textbf{Output Type} & \textbf{Hallucination Type} & \textbf{Continuation Generator} & \textbf{Examples} \\
\midrule
CNN/DM & News article & Bullet-point summary & Fabricated detail & \texttt{gpt-4.1-mini} & 1,000 \\
XSum & BBC article & One-sentence summary & Fabricated detail & \texttt{gpt-4.1-mini} & 1,000 \\
ContraTales & Story prefix & Concluding sentence & Logical contradiction & \texttt{o4-mini} & 2,000 \\
\bottomrule
\end{tabular}
\end{table*}

\paragraph{News Summarisation Datasets} We use CNN/Daily Mail (\textsc{cnn/dm}) \citep{see2017get} and XSum \citep{narayan2018don}, standard abstractive summarisation benchmarks. \textsc{cnn/dm} contains news articles with multi-sentence summaries. XSum features BBC articles with single-sentence summaries, often requiring higher abstraction.

\paragraph{Synthetic Contradictions (ContraTales)} Story prefixes were initially generated using Claude Opus. Then, for these prefixes, factual and hallucinated concluding sentences (outputs) were generated by \texttt{o4-mini}. Hallucinated examples feature a concluding sentence that logically contradicts information established in the prefix (e.g., stating a character who is bald is going for a haircut). This dataset provides unambiguous logical contradictions. See Appendix~\ref{app:contratales_generation} for full details of constructing this dataset.

\paragraph{Data Preparation} For each dataset, we prepared paired examples of a source context and a continuation (output text). Continuations were generated to be either: (i) factual, containing only information directly inferable from the source, or (ii) hallucinated. Hallucinated continuations were produced by prompting \texttt{gpt-4.1} to introduce a plausible but unsupported or contradictory factual detail, while maintaining grammaticality and a sentence length under 40 words. Factual continuations were generated with prompts emphasising strict adherence to the source material, again using \texttt{gpt-4.1}. See Appendix~\ref{app:prompts} for the prompts used to generate continuations.

During evaluation, the source context and its corresponding continuation are concatenated and processed as a single sequence by the observer model.

\subsection{Residual-Stream Linear Probe Methodology}
\label{sec:methods:probe}

\paragraph{Evaluation Protocol}
Unless otherwise specified, all detection methods, including our proposed probe and the baselines, are evaluated using a logistic regression classifier trained to distinguish between factual and hallucinated continuations. Performance is assessed via 5-fold cross-validation. A fixed random seed was used for data splitting and sampling across all experiments to ensure reproducibility.

\paragraph{Residual‑stream linear probe}
Given a frozen observer transformer $\mathcal{F}$ with $L$ decoder blocks and model dimension $d$, let $\mathbf r^{(\ell)}_{t}\in\mathbb{R}^{d}$ denote the post‑layer‑norm residual stream at token position $t$ after block~$\ell$. For each example, we concatenate the source context $X$ and its candidate continuation $Y$, feed the resulting sequence $(x_{0{:}T-1})$ through $\mathcal{F}$, and identify the index of the final token (typically a full stop) of the last sentence in the continuation, $t^{\ast}=T-1$.

From a specific layer $\ell^{\ast}$ (selected via inner‑fold validation on the training set), we extract the activation $\mathbf h = \mathbf r^{(\ell^{\ast})}_{t^{\ast}}$. A logistic probe, parametrised by weights $\mathbf w \in\mathbb{R}^{d}$ and bias $b\in\mathbb{R}$, then predicts the probability of hallucination:
\[
\widehat{y} \;=\; \sigma(\mathbf w^{\top}\mathbf h + b), \quad \text{where} \quad \sigma(z)=\frac{1}{1+e^{-z}}.
\]
The probe is trained using binary cross‑entropy loss with $L_{2}$ regularisation on $\mathbf w$. At test time, the logit $s=\mathbf w^{\top}\mathbf h + b$ serves as the input for a hard decision threshold (typically $s>0$ for hallucination) and as a continuous hallucination score.

\subsection{Baselines}
\label{sec:baselines}
We compare our probe against several baselines. Each baseline generates a feature (or set of features) which is then used to train a logistic regression classifier, following the evaluation protocol described in \S\ref{sec:methods:probe}.

\paragraph{Lexical Overlap} Novelty is quantified as $\nu(Y, X) = 1 - \frac{|\text{n-grams}(Y) \cap \text{n-grams}(X)|}{|\text{n-grams}(Y)|}$ \citep{maynez2020faithfulness}. This is computed for $n \in \{1,2,3\}$, and the maximum score is used as the feature.

\paragraph{Entity Verification} The entity novelty ratio is $\eta(Y, X) = \frac{|E_Y \setminus E_X|}{|E_Y|}$, where $E_X$ and $E_Y$ are sets of named entities extracted from input $X$ and output $Y$ using spaCy \citep{honnibal2020spacy, nan2021entity}.

\paragraph{Semantic Similarity} For the final sentence $s_F$ in the continuation $Y$, we compute its maximum cosine similarity to any sentence in the input $X$: $\phi(s_F, X) = \max_{x_j \in X} \cos(\mathbf{e}(s_F), \mathbf{e}(x_j))$. Sentence embeddings $\mathbf{e}(\cdot)$ are from OpenAI's \texttt{text-embedding-3-small}. The resulting similarity score is used as a feature. (Note: The original paper mentioned flagging segments below a threshold $\tau$; here, we use the score directly as a feature for the logistic regression, which learns the optimal threshold/weighting).

\paragraph{Lookback Lens} Proposed by \citet{Chuang2024LookbackLens}, this method analyses attention patterns. Given concatenated context $X$ (length $N$) and continuation $Y$, for each head $(l,h)$ at position $t$ in $Y$, it computes average attention to context tokens $A_t^{l,h}(\text{context})$ and to preceding tokens in $Y$, $A_t^{l,h}(\text{new})$. The lookback ratio is $\text{LR}_t^{l,h} = \frac{A_t^{l,h}(\text{context})}{A_t^{l,h}(\text{context}) + A_t^{l,h}(\text{new})}$. For the final sentence of $Y$, these ratios are averaged over its tokens for each head, forming a feature vector $\overline{\mathbf{v}}$ by concatenating these values across all heads and layers. This vector $\overline{\mathbf{v}}$ is input to the logistic regression classifier. Our observer paradigm processes the concatenated text; $N$ is the length of the original source context $X$.

\subsection{Unit‑Level Hallucination Attribution}
\label{sec:methods:unit_attr}

To identify transformer sub‑modules influencing the hallucination score, we use gradient‑times‑activation.
\paragraph{Notation.}
Let the probe score for a sequence, derived from the optimal layer $\ell^{\ast}$ and final token $t^{\ast}$ as defined in \S\ref{sec:methods:probe}, be $s=\mathbf w^{\top}\mathbf r^{(\ell^{\ast})}_{t^{\ast}} + b$. For any layer $\ell\in\{0,\dots,L\!-\!1\}$ and token position $t$:
\begin{align*}
\mathbf r^{(\ell)}_{t} &\in\mathbb{R}^{d}\quad &&\text{post‑LN residual stream,} \\
\mathbf z^{(\ell,h)}_{t} &\in\mathbb{R}^{d}\quad &&\text{output of head $h$ in layer $\ell$ (after $W^{O}$),}\\
\mathbf m^{(\ell)}_{t} &\in\mathbb{R}^{d}\quad &&\text{output of the MLP in layer $\ell$ (after $W^{\text{out}}$).}
\end{align*}

\paragraph{Gradient–times–activation.}
We compute gradients of the probe score $s$ with respect to intermediate residual stream activations: $\mathbf g^{(\ell)}_{t} = \partial s / \partial \mathbf r^{(\ell)}_{t}$. The contribution $a_{t}(\mathbf u)$ of a module's output vector $\mathbf u_{t}$ (where $\mathbf u_{t} \in \{\mathbf z^{(\ell,h)}_{t}, \mathbf m^{(\ell)}_{t}\}$ which contributes to a subsequent residual stream $\mathbf r^{(\ell')}_{t}$) is taken as its projection onto the gradient of that residual stream: $a_{t}(\mathbf u) = \langle \mathbf g^{(\ell')}_{t}, \mathbf u_{t} \rangle$. This value indicates if the module's output nudges the relevant residual stream in the direction that increases (positive) or decreases (negative) the hallucination score $s$. These contributions are averaged over the tokens $\mathcal{S}$ of the final sentence in the continuation:
\begin{align*}
A^{(\ell,h)}_{\text{head}} &= \frac{1}{|\mathcal{S}|} \textstyle\sum_{t\in\mathcal{S}} a_{t}(\mathbf z^{(\ell,h)}) \\ A^{(\ell)}_{\text{mlp}} &= \frac{1}{|\mathcal{S}|} \textstyle\sum_{t\in\mathcal{S}} a_{t}(\mathbf m^{(\ell)}).
\end{align*}

\paragraph{Dataset‑level aggregation.}
To analyse general trends, we compute these attributions for the $N=100$ examples with the highest hallucination scores (i.e., most confidently predicted as hallucinations by the probe) from each dataset and report the mean $\bar{A} = \frac{1}{N}\sum_{n=1}^{N}A_{(n)}$. This focuses the analysis on mechanisms related to strong hallucination signals. Such gradient-based attributions offer first-order estimates of causal influence: rescaling a unit’s output by $(1+\delta)$ would be expected to shift the probe logit by approximately $\delta\bar{A}$.

\section{Results}
\label{sec:results}

\subsection{Language models exhibit a robust \emph{linear} representation of contextual hallucinations}
\label{sec:results:linear_rep}

\begin{figure*}[t]
    \centering
    \includegraphics[width=0.85\linewidth]{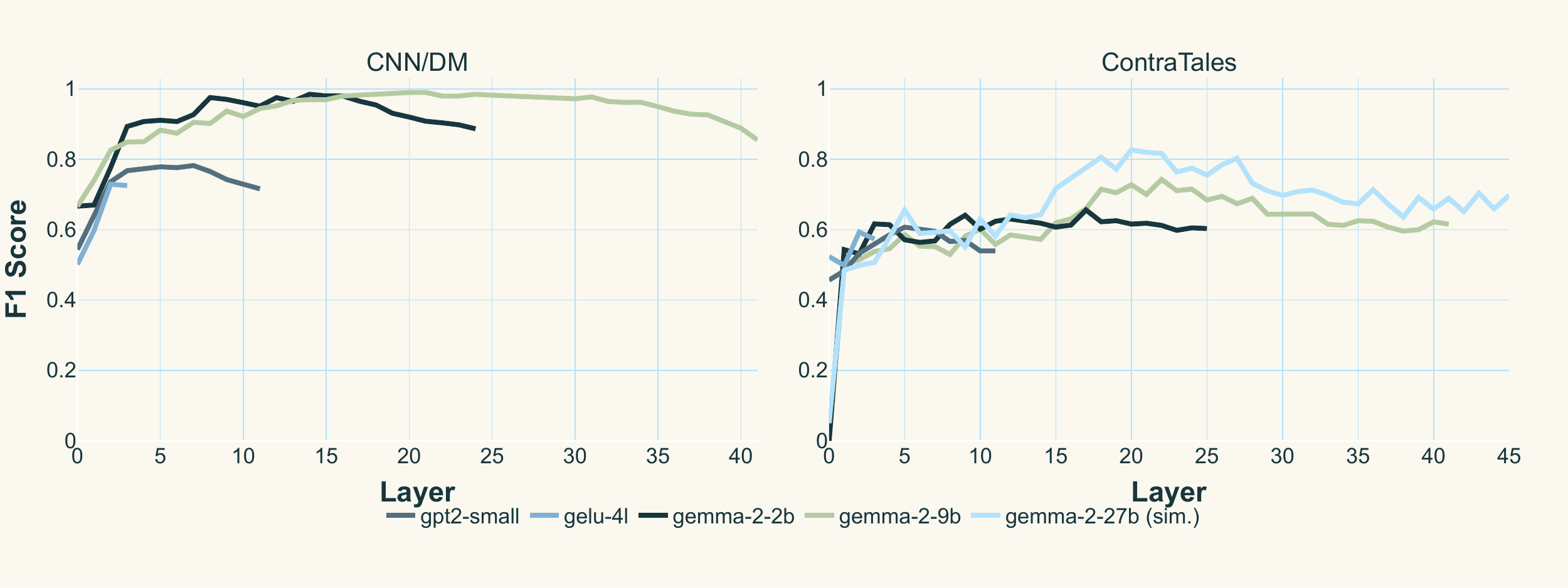}
    \caption{\textbf{Layer‑wise detection performance of residual‑stream linear probes.}  
    Each curve shows the F\textsubscript{1} score (5‑fold CV) of a logistic probe trained on a single transformer layer to classify the final sentence of a document as hallucinated or supported by context. \textbf{Left}: results on \textsc{cnn/dm} summarisation; \textbf{right}: results on the synthetic‐contradiction \textsc{ContraTales}.  The consistent mid‑layer plateau across four observer models supports the hypothesis that contextual hallucinations are encoded along a common linear direction in activation space.}
    \label{fig:combined_layer_sweep}
\end{figure*}

To test the linear representation hypothesis—that a single direction in residual-stream activation space separates hallucinated from supported spans—we trained linear probes on this space. Figure~\ref{fig:combined_layer_sweep} presents the F\textsubscript{1} scores of logistic regression probes, trained on each layer of Gemma‑2 (2B, 9B, 27B), GPT‑2-small, and a 4‑layer GELU baseline.\footnote{\texttt{gelu-4l} in TransformerLens: \url{https://transformerlensorg.github.io/TransformerLens/generated/model_properties_table.html}} Evaluations were performed on \textsc{cnn/dm} news summaries and the \textsc{ContraTales} synthetic contradiction dataset.

Probe performance typically rises in early layers, peaks in mid-to-late transformer blocks, and then plateaus. On \textsc{cnn/dm}, Gemma-2-9B achieved an F\textsubscript{1} of $0.98$ by layer 17, maintaining $>0.95$ through layer 37. GPT‑2-small reached $0.78$ at layer 11. On \textsc{ContraTales}, Gemma-2-9B reached $0.70$ in its best layers, and Gemma-2-27B achieved $0.84$. The consistent mid-layer performance plateau across models supports a single-direction explanation. This direction generally emerges by layers 8–12, with deeper layers offering marginal improvement in discriminatory power. The optimal layer for detection tends to be deeper in models with more parameters.

The extent to which these probes exploit superficial lexical cues is addressed by comparison to baselines in Section~\ref{subsec:outperform}. The uniqueness of this linear direction is not established here, nor is causality, which is investigated in Section~\ref{sec:results:steering} through generation steering.

\subsection{Residual‑stream probes outperform heuristic detectors}
\label{subsec:outperform}

\begin{figure*}[t]
    \centering
    \includegraphics[width=0.8\linewidth]{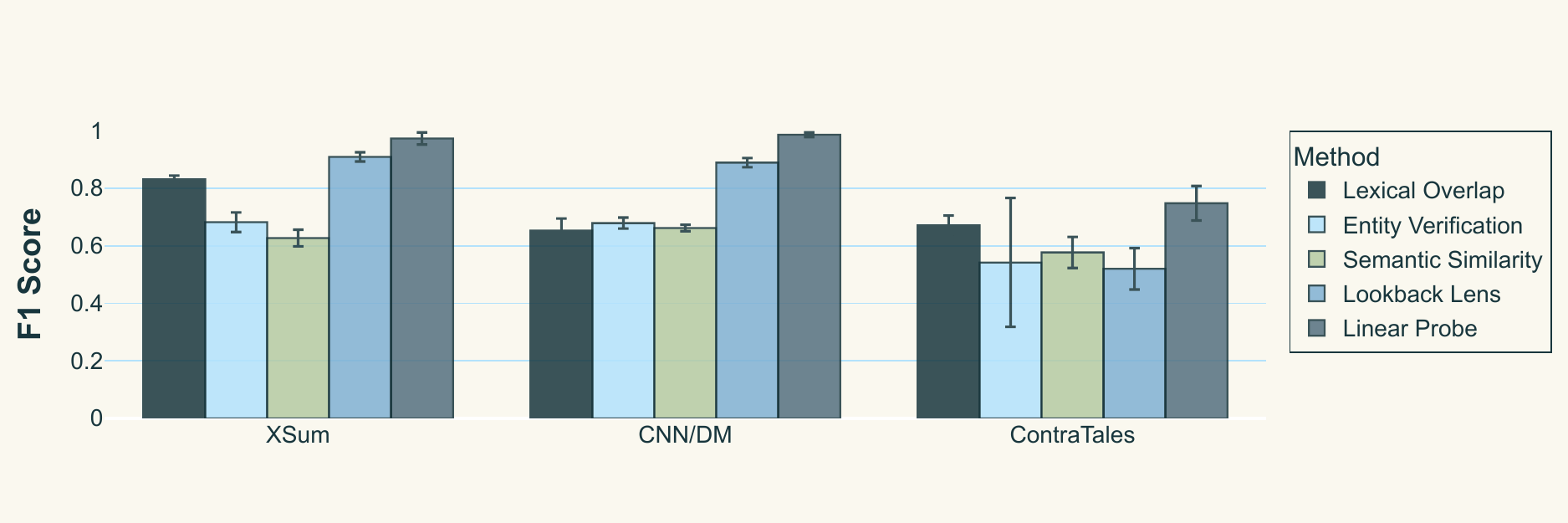}
    \caption{\textbf{Comparison of hallucination‑detection methods.}  
    Bars give mean F\textsubscript{1} over five cross‑validation folds; whiskers show the $95\%$ bootstrap confidence interval.  The residual‑stream \emph{linear probe} (right‑most bar in each group) consistently exceeds all baselines -- lexical overlap, entity verification, semantic similarity, and Lookback~Lens -- across the news datasets (\textsc{xsum}, \textsc{cnn/dm}) and the logically harder \textsc{ContraTales}.}
    \label{fig:method_sweep}
\end{figure*}

Figure~\ref{fig:method_sweep} shows F\textsubscript{1} scores for the residual-stream linear probe against four baseline detectors -- lexical overlap, entity verification, semantic similarity, and Lookback Lens -- across three datasets. On news benchmarks, the linear probe achieved $0.97\pm0.01$ F\textsubscript{1} on \textsc{xsum} and $0.99\pm0.01$ on \textsc{cnn/dm}. This surpassed the strongest baseline, Lookback Lens, by 5–8 points and lexical measures by approximately 15 points.

On the \textsc{ContraTales} dataset, the performance gap increased. Lexical overlap and entity verification F\textsubscript{1} scores were $0.66$ and $0.55$, respectively. Lookback Lens scored $0.48\pm0.11$. The linear probe achieved $0.75\pm0.04$ F\textsubscript{1}, outperforming these alternatives by 9–27 points. This dataset features contradictions that are primarily logical rather than lexically obvious, a characteristic that diminished the effectiveness of the baseline methods which target surface-level cues or specific attention shifts. The linear probe's F\textsubscript{1} score on \textsc{ContraTales}, while higher than baselines, did not reach its news-domain performance levels.

\subsection{Hallucination representation transfers between news datasets}

\begin{figure*}
\centering
\includegraphics[width=0.8\linewidth]{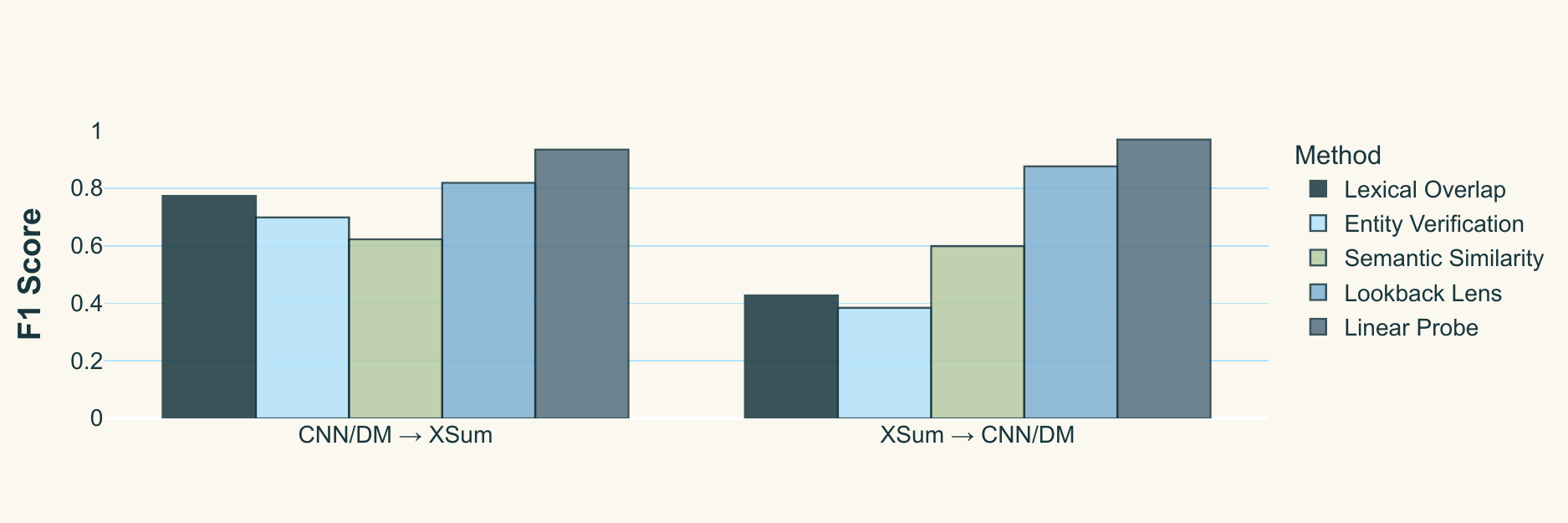}
    \caption{\textbf{Cross-domain transfer performance of hallucination detection methods.} F\textsubscript{1} scores for detectors trained on one news dataset (\textsc{cnn/dm} or \textsc{xsum}) and evaluated on the other. Features were extracted from layer 20 of a Gemma-2-9B observer. The linear probe demonstrates high transferability compared to baseline methods.}
\label{fig:transfer_sweep}
\end{figure*}

\paragraph{Cross-domain news transfer}
Figure~\ref{fig:transfer_sweep} shows cross-domain generalisation performance. Detectors were trained on one news corpus (\textsc{cnn/dm} or \textsc{xsum}) and evaluated on the other without re-tuning. All features were extracted from layer 20 of a Gemma-2-9B observer model. The linear probe exhibited minimal accuracy loss from domain shift. Lookback Lens also generalised to an extent. In contrast, surface cue-based methods showed substantial performance drops. For instance, lexical overlap F\textsubscript{1} decreased from $0.78$ (in-domain \textsc{cnn/dm}) to $0.42$ when transferred to \textsc{xsum}. Semantic similarity performance was near chance levels out-of-domain. 

\begin{figure}
\centering
\includegraphics[width=0.7\linewidth]{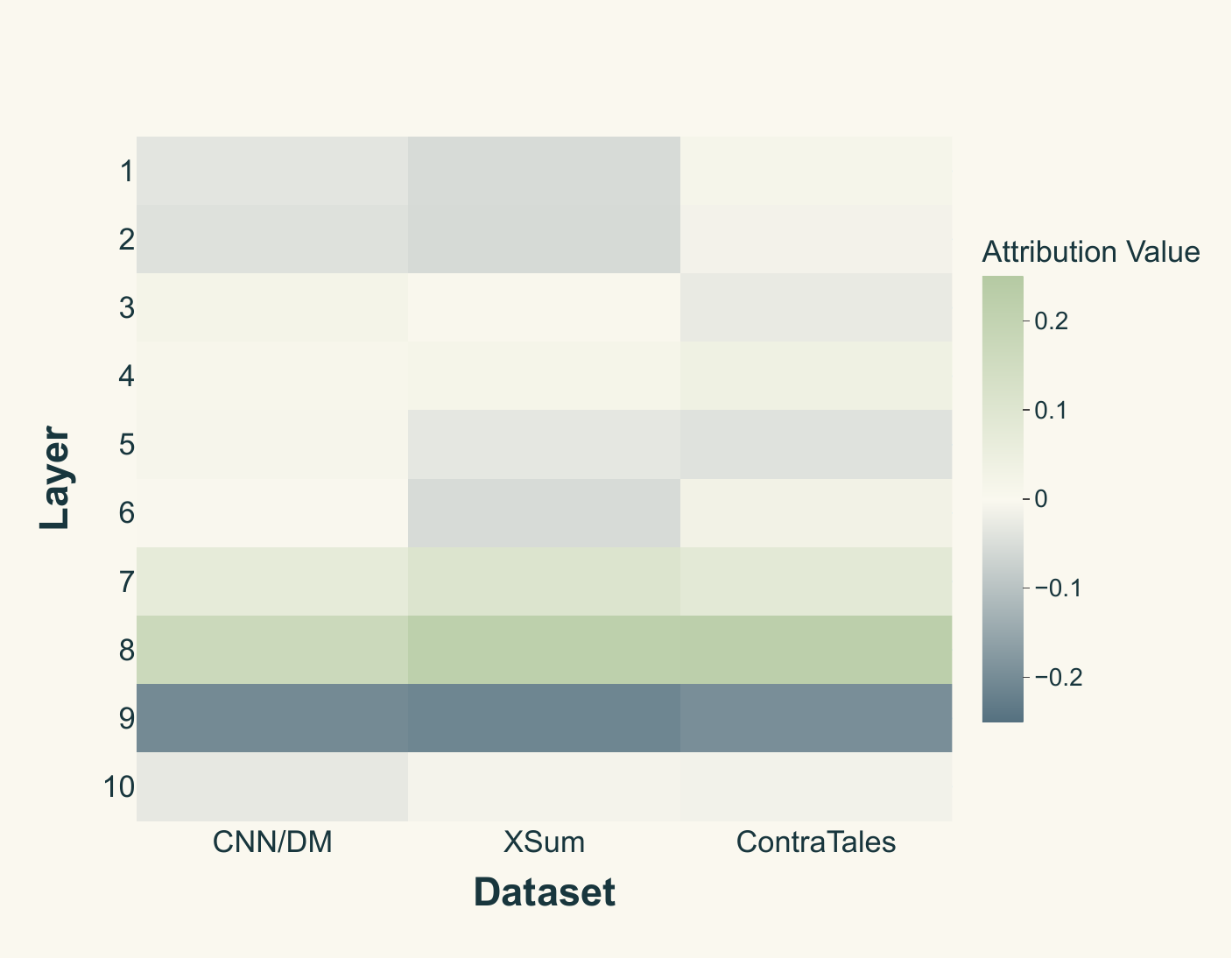}
    \caption{\textbf{Aggregated MLP layer attributions for the hallucination probe.} Mean MLP attributions ($\bar{A}^{(\ell)}_{\text{mlp}}$) per layer for a linear probe trained on layer 10 of Gemma-2-9B. Attributions are presented for evaluations on \textsc{cnn/dm}, \textsc{xsum}, and \textsc{ContraTales}, revealing a consistent pattern across datasets in layers 7-9.}
\label{fig:mlp_attributions_heatmap}
\end{figure}

\paragraph{MLP attributions identify a consistent sub-circuit}
Figure~\ref{fig:mlp_attributions_heatmap} presents aggregated MLP attributions, $\bar{A}^{(\ell)}_{\text{mlp}}$, for a linear probe trained on layer 10 of Gemma-2-9B and evaluated on \textsc{cnn/dm}, \textsc{xsum}, and \textsc{ContraTales}.

Per-head attention attributions showed fluctuations around zero, lacking layer-consistent signs or overlap in top-ranked heads across datasets. This indicates that attention routing, in this observer setup, does not offer a stable attribution signal for contextual hallucination.

In contrast, MLP attributions were sparse and layer-consistent. For the layer 10 probe, layers 7 and 8 exhibited positive attribution towards the hallucination direction (with layer 8 being dominant), while layer 9 showed a strong negative attribution. Contributions from other layers were minimal ($|\bar{A}| < 0.02$). This specific \{positive (layer 7) $\to$ strong-positive (layer 8) $\to$ negative (layer 9)\} MLP attribution pattern was consistently observed across all three datasets.

\paragraph{Probe Activations on Unrelated Text}
The hallucination probe trained on Gemma-2-9B was applied to one million 256-token sequences from the Pile \citep{gao2020pile}. Analysis of the sequences producing the highest and lowest probe activations revealed distinct patterns. No consistent pattern was identified in the highest-activating examples. However, the lowest-activating examples, detailed in Appendix~\ref{app:pile} (Table~\ref{tab:pile}), consistently featured textual repetition. This included exact phrase repetitions (e.g., in gaming or medical texts), quoted content (e.g., from forums or chat logs), and formulaic language found in technical or religious documents. The strongest negative activations (e.g., around -30) were associated with such repeated content.

\subsection{Hallucination representation can be used to steer generation}
\label{sec:results:steering}

\begin{figure*}
    \begin{minipage}{0.48\textwidth}
        \centering
        \includegraphics[width=0.98\linewidth]{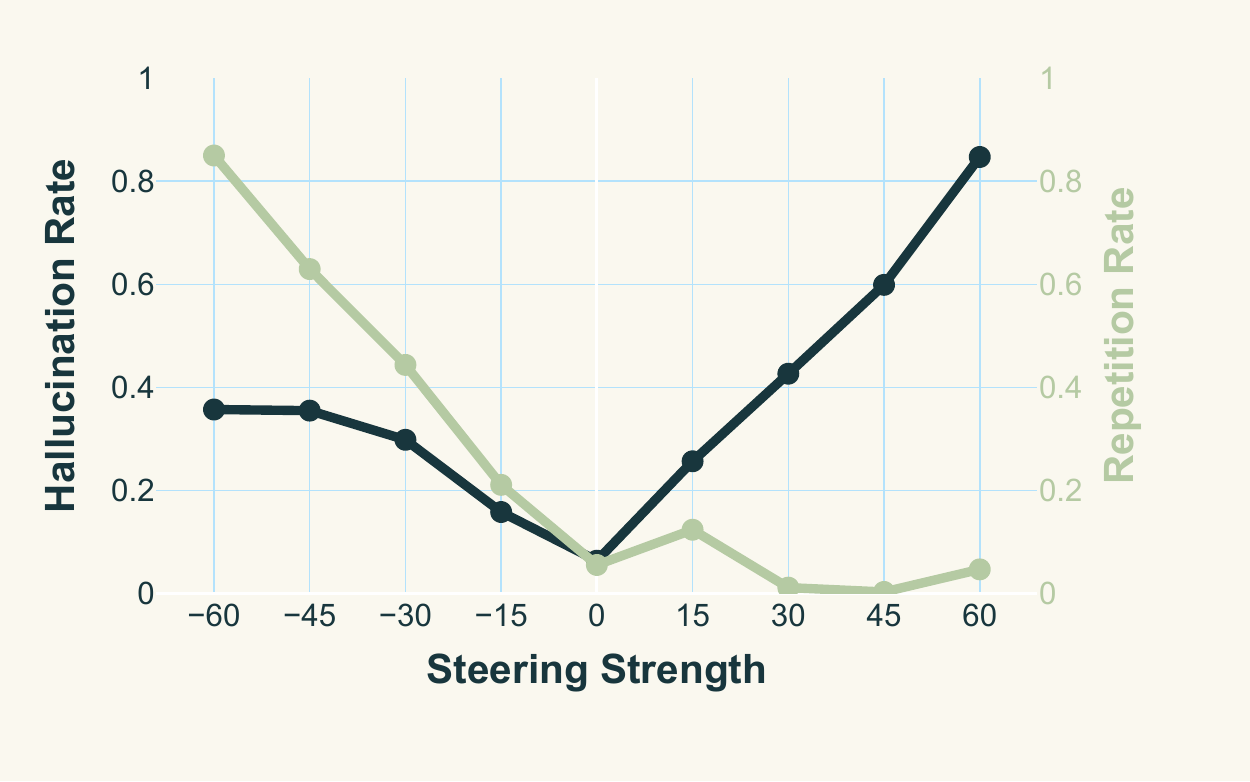}
        \caption{We use steering to generate outputs for CNNDM with the Gemma-2-2B model. We then use fuzzy string matching to determine the repetition rate, and gpt-4.1 to determine the hallucination rate.}
        \label{fig:steering_effects}
    \end{minipage}
    \hfill
    \begin{minipage}{0.48\textwidth}
        \centering
        \includegraphics[width=\linewidth]{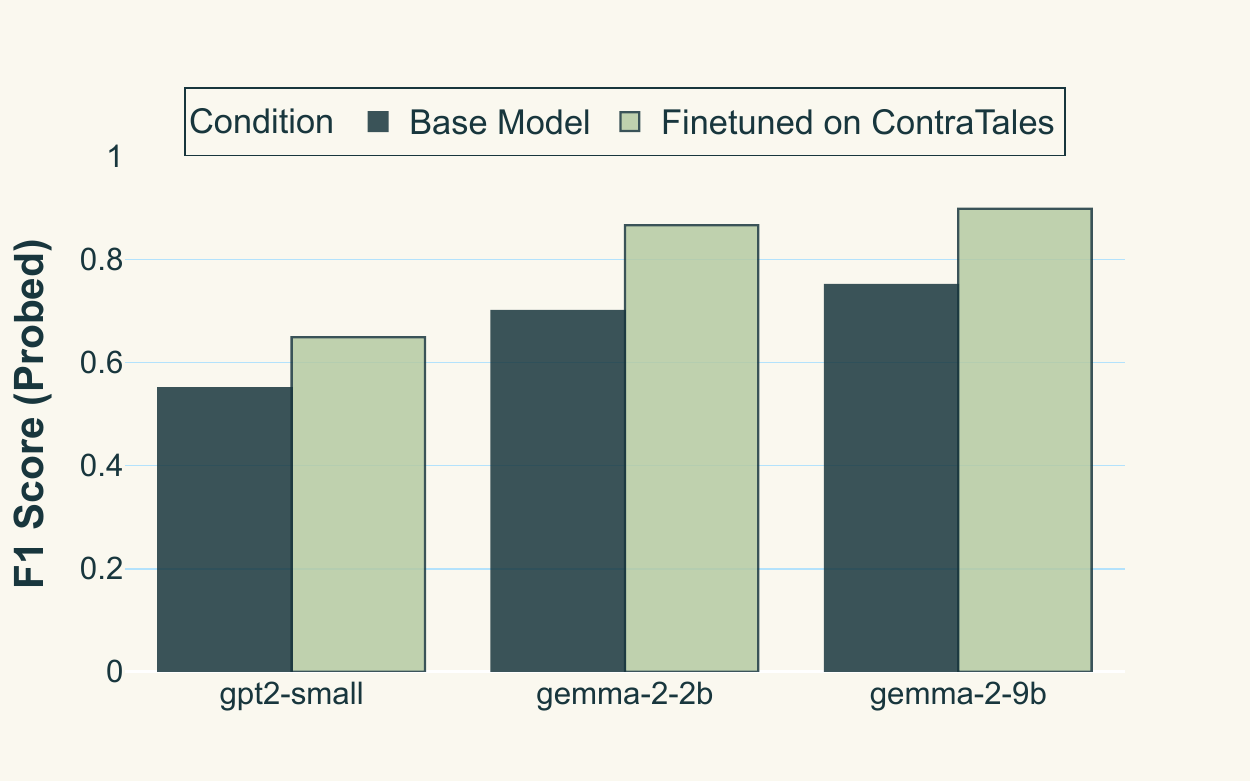}
        \caption{Unsupervised domain adaptation boosts probe accuracy. Bars report the F\textsubscript{1} of the residual‑stream logistic probe before (base) and after (FT) two‑epoch SFT on 1000 \emph{correct} \textsc{ContraTales} continuations.}
        \label{fig:finetuning_effects}
    \end{minipage}
\end{figure*}

To test the causal effect of the identified residual-stream direction, we patched the normalised probe vector $\mathbf{w}/\!\|\mathbf{w}\|$ (derived from a linear probe on layer 10 activations of the \emph{same Gemma-2-2B model architecture} used for generation) into its layer 10 during \textsc{CNN/DailyMail} summarization (50 new tokens, greedy decoding). This vector, specifically trained to distinguish hallucinated from faithful content for the Gemma-2-2B, was scaled by $\alpha \in \{-60, \dots, +60\}$ and injected once at generation start. We generated 128 summaries per $\alpha$, measuring hallucination rate (judged by GPT-4.1, prompt in App.~\ref{app:prompts}) and repetition rate (RapidFuzz, $\ge 5$-grams, ratio $>85\%$). The results, plotted in Fig.~\ref{fig:steering_effects}, demonstrate bidirectional control: positive scaling (e.g., $\alpha=+60$) increased hallucination to $0.86$ while reducing repetition below $0.05$, whereas negative scaling (e.g., $\alpha=-60$) increased repetition to $0.84$ with a hallucination rate of $0.35$. The unpatched model ($\alpha=0$) served as a baseline for this trade-off.

\subsection{Finetuning improves internal hallucination indication}
\label{sec:results:finetune}

To investigate improving observer performance on a target domain without new hallucination labels, we performed unsupervised domain adaptation. Observer models (GPT-2-small, Gemma-2-2B, Gemma-2-9B) were further trained for two epochs on 1000 correct-only completions from the \textsc{ContraTales} corpus using standard SFT hyperparameters (AdamW, LR $1 \times 10^{-5}$, context length 512, batch size 8, 8xH200 GPUs, no dropout). Following this adaptation, the logistic residual-stream probe (as per \S\ref{sec:methods}) was retrained on the original labeled data and evaluated on an unseen test fold. As shown in Figure~\ref{fig:finetuning_effects}, this process improved F\textsubscript{1} scores for all models: by +0.10 for GPT-2-small, +0.17 for Gemma-2-2B, and +0.14 for Gemma-2-9B. For Gemma-2-9B on \textsc{ContraTales}, the F\textsubscript{1} score increased from $0.75$ to $0.89$.

\section{Discussion}

This work demonstrates a practical and actionable application of interpretability insights through a generator-agnostic observer paradigm: a linear probe on a transformer's residual-stream activations identifies contextual hallucinations in a single forward pass, achieving high F\textsubscript{1} scores. These results underscore the feasibility of leveraging internal model representations for addressing key AI challenges like hallucinations. While the F\textsubscript{1} scores on news benchmarks could be partially influenced by characteristics inherent in prompted synthetic hallucinations, the strong performance on \textsc{ContraTales} (a dataset designed to test unambiguous logical contradictions) mitigates this concern by showcasing the probe's ability to identify more fundamental contextual violations, arguably less susceptible to specific generation artifacts.

Our findings offer support for the linear representation hypothesis (LRH) in contextual understanding, showing how interpretability can move beyond correlation to causal intervention. The identified linear direction for hallucination's consistency across Gemma-2 model sizes (2B$\to$9B) and transferability across news domains suggest a fundamental encoding. Crucially, its functional role is substantiated by causal interventions: a single, layer-local injection or ablation of the probe vector along this axis smoothly and monotonically modulates a generator's hallucination and repetition rates, demonstrating it as an actionable, low-dimensional, and causally effective axis for contextual-hallucination awareness. Mechanistically, gradient-times-activation attribution analyses refine this picture, pinpointing the signal for contextual inconsistency to a sparse, layer-consistent pattern of late-layer MLP activity (e.g., layers 7 and 8 positively, layer 9 negatively, for a probe on layer 10 of Gemma-2-9B), rather than diffuse attention patterns. This indicates the observer’s awareness is canalised through a specific chain of late‑layer feed‑forward computations.

Furthermore, the practical utility of this internal representation is practically enhanced by unsupervised domain adaptation. Finetuning the observer model on in-domain, correct-only text from \textsc{ContraTales} (without new hallucination labels) significantly improved the logistic probe's F\textsubscript{1} score (e.g., for Gemma-2-9B, from 0.75 to 0.89). This implies that additional in‑domain language modelling sharpens the internal distinction between logically supported and unsupported statements, a distinction the linear probe can then more effectively exploit, offering an inexpensive, label‑free path to enhanced single‑pass detection capabilities: important for real-world deployment.

\paragraph{Limitations} A primary limitation of this work is the reliance on synthetically generated hallucinations for training and evaluating the majority of our detectors, particularly on the news and medical datasets. While necessary for creating labeled data at scale, continuations prompted from large language models may exhibit patterns or artifacts predictable to an observer model trained on similar data, which might not be representative of naturally occurring ``in-the-wild'' hallucinations generated by various models under different conditions. This could potentially lead to an overestimation of the detector's performance and generalisability beyond the specific generation methods used here. Although the \textsc{ContraTales} dataset offers a valuable benchmark for pure logical contradictions, it also represents a specific type of structured inconsistency. Further validation on datasets containing a broader spectrum of organically generated, human-verified hallucinations would strengthen claims regarding real-world applicability.

Beyond the nature of the training data, the evaluation of the steering experiments relies on a \texttt{gpt-4.1} judge to determine hallucination rates. While this is a common method, LLM-based evaluations can be subject to noise, bias, and variability, which may affect the precision of the quantitative steering results. Additionally, while the linear probe detector itself is lightweight and efficient, the observer paradigm necessitates deploying and running a sufficiently capable base transformer model, which still carries significant computational costs compared to purely surface-level detection heuristics. Finally, our study focuses specifically on \textbf{intrinsic} hallucinations that contradict or are unsupported by the provided source context; the applicability of the discovered hallucination direction and detection method to \textbf{extrinsic} hallucinations, which introduce novel but unverifiable information from outside sources, remains untested.

This research provides compelling evidence for a single, transferable, and causally effective linear direction within transformer activations that corresponds to contextual hallucination. This direction is primarily processed by a sparse and consistent MLP sub-circuit and can be leveraged for both lightweight detection and controlled generation, advancing interpretability by providing concrete methods for building more reliable AI systems. To facilitate further research, we release the \textsc{ContraTales} benchmark.

\section*{Impact Statement}

This paper advances mechanistic interpretability by improving AI reliability and safety through evidence of generator-agnostic methods for detecting and controlling contextual hallucinations via mechanistic insights and causal steering. This offers potential for positive societal impact, enabling the deployment of more trustworthy AI in domains where accuracy is essential. However, this work also carries potential risks; the steering technique, while intended for mitigation, could be misused for generating misinformation, and the detector's performance may be influenced by biases present in its training data.

\bibliography{example_paper}
\bibliographystyle{icml2025}

\newpage
\appendix
\onecolumn

\section{Generation Prompts}
\label{app:prompts}

This section details the prompts used to generate hallucinated and factual continuations for our datasets. We used the OpenAI API with GPT-4.1-mini to create both hallucinated and factual continuations by providing specific instructions through system and user messages.

\subsection{Hallucination Generation Prompts}

For generating hallucinated continuations (introducing unsupported or contradictory information), we used the following system and user prompts:

\begin{lstlisting}[backgroundcolor=\color{lightest_blue}, frame=single, basicstyle=\ttfamily\small, breaklines=true]
# System Message
You are an expert summarization assistant helping to create a hallucination dataset. 
Your task is to produce exactly ONE sentence that *appears* relevant to the given article but introduces at least one factual detail that cannot be inferred from the article.
The sentence must be grammatically correct, and <= 40 words.

# User Message
Original article:
<article>
{document}
</article>

Original correct summary:
<summary>
{original_summary}
</summary>

We are going to replace this sentence: <sentence_to_replace>{replaced_sentence}</sentence_to_replace> with a hallucinated sentence that you generate.

The new sentence must contain a made up factual detail that is not present in the original article.

Return JUST the new sentence, without quotation marks, xml tags, or any additional commentary.
\end{lstlisting}

\subsection{Factual Generation Prompts}

For generating factual continuations (containing only information supported by the source), we used the following system and user prompts:

\begin{lstlisting}[backgroundcolor=\color{lightest_blue}, frame=single, basicstyle=\ttfamily\small, breaklines=true]
# System Message
You are an expert summarization assistant helping to create a dataset.
You will be given an existing summary and asked to replace a single sentence with one written by you.
The sentence you write must be factually accurate, grammatically correct, and only contain information directly from the original article.
Your sentence should be concise and <= 40 words.

# User Message
Original article:
<article>
{document}
</article>

Original correct summary:
<summary>
{original_summary}
</summary>

We are going to replace this sentence: <sentence_to_replace>{replaced_sentence}</sentence_to_replace> with a new sentence that you generate.

The new sentence must be completely factual, only containing information directly from the original article.

Return JUST the new sentence, without quotation marks, xml tags, or any additional commentary.
\end{lstlisting}

\subsection{Hallucination Evaluation Prompt}

For evaluating hallucination rates during the steering experiments in Section \ref{sec:results:steering}, we used GPT-4.1 with the following prompt:

\begin{lstlisting}[backgroundcolor=\color{lightest_blue}, frame=single, basicstyle=\ttfamily\small, breaklines=true]
You are an expert at detecting hallucinations in summaries.

I will provide you with an article and a summary. Your task is to determine if the summary contains any hallucinations - information that is not supported by or contradicts the article.

A hallucination in a summary is when:
1. It contains information NOT present in the article
2. It contradicts information in the article
3. It makes claims that require outside knowledge not in the article

Answer with ONLY "hallucination" or "no hallucination".

Article:
{article}

Summary:
{summary}
\end{lstlisting}

These prompts were implemented in our data generation pipeline to create paired examples of source contexts and either factual or hallucinated continuations across all datasets described in Section \ref{sec:datasets}.

\section{\textsc{ContraTales} Dataset Generation}
\label{app:contratales_generation}

The \textsc{ContraTales} dataset, comprising 2000 examples of stories with purely logical contradictions, was generated to test the contextual understanding of hallucination detectors. The generation process involved using a large language model, specifically \texttt{o4-mini}, guided by a detailed prompt that included instructional guidelines and few-shot examples.

\subsection{Generation Process}

Each example in \textsc{ContraTales} consists of three main parts:
\begin{enumerate}
    \item \textbf{Story Prefix}: An initial narrative segment (typically 7-10 sentences). The first sentence of this prefix establishes an unambiguous constraint or fact about a character or situation (e.g., ``Jack had been bald for 10 years,'' ``Sarah was allergic to peanuts''). The subsequent sentences in the prefix develop a neutral, everyday scene, intentionally avoiding any direct reference to, or negation of, the established constraint.
    \item \textbf{Correct Concluding Sentence}: A single sentence that provides a logical and coherent continuation of the story prefix, respecting the initial constraint.
    \item \textbf{Contradictory Concluding Sentence}: A single sentence that is structurally and tonally similar to the correct concluding sentence but introduces a detail that subtly and logically contradicts the constraint established in the first sentence of the story prefix.
\end{enumerate}

The generation was performed using the \texttt{o4-mini} model. The model was prompted with a set of instructions that specified the desired structure and characteristics of the \textsc{ContraTales} examples. To guide the model effectively, the prompt also included a random selection of 5 few-shot examples from a seed set of 24 pre-existing \textsc{ContraTales}. These examples demonstrated the desired format and subtlety. The aim was to create contradictions that are primarily logical rather than lexically obvious, thereby challenging detection methods reliant on surface-level cues.

\subsection{Example \textsc{ContraTales}}

Table \ref{tab:contratales_examples} showcases four examples from the \textsc{ContraTales} dataset, illustrating the structure and nature of the logical contradictions.

\begin{table}[H]
\centering
\caption{Illustrative examples from the \textsc{ContraTales} dataset.}
\label{tab:contratales_examples}
\resizebox{\textwidth}{!}{%
\begin{tabular}{@{}p{0.4\textwidth}p{0.27\textwidth}p{0.27\textwidth}@{}}
\toprule
\textbf{Story Prefix (Edited for brevity)} & \textbf{Correct Concluding Sentence} & \textbf{Contradictory Concluding Sentence} \\
\midrule
Jack had been bald for 10 years. Each morning, he stepped onto his small porch to water the potted herbs... By mid-morning, Jack was usually cycling through the neighborhood... Later, he settled on a bench in the park... In the afternoons, he volunteered at a literacy program... Tonight, he planned to check out the grand opening of an artisan soap store nearby. & Jack was going to the barber shop to get scalp oil. & Jack was going to the barber shop to get a haircut. \\
\midrule
Sarah was allergic to peanuts. On Saturday morning, she and her friend packed their bags for a scenic hike... Sarah filled her water bottle and double-checked her map... At a shaded clearing, they paused... Sarah pulled out a granola bar from her pack... Just before dusk, they reached a picnic table overlooking the valley below. & Sarah refused the PB\&J sandwich her friend offered. & Sarah ate the PB\&J sandwich her friend offered. \\
\midrule
Daniel can't swim. He spent Saturday mornings at the community center's lounge reading magazines... Daniel often joined the adult painting sessions instead of the pool activities... Afterwards, he sketched scenes from nature in his notebook. He left the lounge and walked toward the pool deck. There, he paused at the edge of the water, watching the ripples in the sunlight. & Daniel watched the swimmers from a bench by the pool. & Daniel swam laps to warm up before the race. \\
\midrule
Laura didn't own a smartphone. Each day she followed her paper planner to keep track of appointments... At the office, she used the wall calendar to schedule meetings... During breaks, she called home from the lobby phone... For lunch, she read a paperback novel... Before dinner, she walked to the hotel and paused under its marquee. & Laura checked the directory in the lobby to find the restaurant's address. & Laura sent a text to reserve a table. \\
\bottomrule
\end{tabular}%
}
\end{table}

\section{Manual Features}
\label{app:manual_features}

\subsection{Feature extraction}

Let $X=(x_{1},\dots ,x_{|X|})$ be the context and $Y=(y_{1},\dots ,y_{|Y|})$ the model-generated continuation.
We concatenate them and perform one forward pass through the 40-layer Gemma-2-27B model $f_{\theta}$, storing the logits $L\in\mathbb{R}^{T\times V}$, the attention patterns
$A^{(\ell)}\in[0,1]^{T\times T}$ for layers $\ell\in\{40,42,44\}$ (softmaxed over keys) and the mid-layer residual stream
$R^{(m)}\in\mathbb{R}^{T\times d_{model}}$ at layer $m=28$.
The continuation is segmented into non-overlapping chunks $\mathcal{C}_{Y}=\{c^{Y}_{1},\dots ,c^{Y}_{n}\}$ by splitting on full stops and newlines; the context is chunked analogously into $\mathcal{C}_{X}=\{c^{X}_{1},\dots ,c^{X}_{k}\}$.

For each note chunk $c^{Y}_{i}$ we derive a dense feature vector $\phi(c^{Y}_{i})$ composed of nine orthogonal signal families described below; all computations reuse tensors already cached, thereby avoiding additional forward passes or temperature sampling.

\paragraph{Token-level uncertainty}
Given the gold tokens $y_{t}\in\mathbb{N}^{V}$ inside $c^{Y}_{i}$ we compute the negative log-likelihood $\operatorname{NLL}{t}=-\log p_{\theta}(y_{t})$ and the token entropy $H_{t}=-\sum_{v}p_{\theta}(v)\log p_{\theta}(v)$. We record the mean and maximum NLL, the proportion of tokens with ground-truth rank $>10$ and the slope of a least-squares fit of $H_{t}$ versus token position.

\paragraph{Cross-context attention}
Let $\bar{a}^{(\ell)}{Y\to X}(i,j)$ be the mean of $A^{(\ell)}$ over all query positions in chunk $c^{Y}_{i}$ and key positions in $c^{X}_{j}$; similarly $\bar{a}^{(\ell)}{Y\to Y}(i)$ averages over keys in $c^{Y}_{i}$ itself. We form the self-ratio $s^{(\ell)}{i}=\bar{a}^{(\ell)}{Y\to Y}(i)\big/\bigl(\frac{1}{k}\sum{j}\bar{a}^{(\ell)}{Y\to X}(i,j)+10^{-6}\bigr)$ and summary statistics of the top-15 values $\{\bar{a}^{(\ell)}{Y\to X}(i,j)\}_{j}$.

\paragraph{Residual-stream semantic alignment}
Using $\tilde{r}_{i}= \| \sum{t\in c^{Y}_{i}}R^{(m)}_{t}\|^{-1}_{2}\sum{t\in c^{Y}_{i}}R^{(m)}_{t}$ we measure its cosine similarity to (i) the residual average of the transcript chunk with maximal attention weight and (ii) the average residual of the entire transcript.

\paragraph{Embedding-based similarity}
Chunks are embedded with the OpenAI \texttt{text-embedding-3-small} model, yielding unit-norm vectors $\{e^{Y}_{i}\}$ and $\{e^{X}_{j}\}$. We store the maximum, mean top-3 and variance of $\langle e^{Y}_{i},e^{X}_{j}\rangle$ across $j$, together with the gap between the two highest scores and global max/mean similarities.

\paragraph{Logit eigenspectrum}
For the slice $L_{c^{Y}_{i}}\in\mathbb{R}^{|c^{Y}_{i}|\times V}$ we take the top-32 singular values $\sigma_{1}\ge\dots\ge\sigma_{32}$ and include $\sum_{r}\log\sigma_{r}$ and the spectral gap $\sigma_{1}-\sigma_{2}$.

\paragraph{Entity grounding}
Clinical entities are extracted using SciSpaCy.
We compute the coverage ratio $|E_{i}\cap E_{X}|/|E_{i}|$ and the count of novel entities $|E_{i}\setminus E_{X}|$, where $E_{i}$ and $E_{X}$ are the entity sets of chunk $i$ and the transcript respectively.

\paragraph{Surface-level heuristics}
Features include trigram novelty -- $|\operatorname{Tri}(c^{Y}_{i})\setminus\operatorname{Tri}(X)|\big/|\operatorname{Tri}(c^{Y}_{i})|$ -- numeric-token ratio and the z-score of mean sentence length.

\paragraph{Intra-chunk semantic-graph variance}
A $k\!\!=\!3$ cosine k-NN graph is built on sentence-level embeddings within the chunk; the feature is the variance of edge similarities, capturing semantic isolation.

\begin{table}[htbp]
  \centering
  \caption{Chunk-level feature families used for hallucination classification. All quantities are derived from a single forward pass or from cached embeddings.}
  \label{tab:features}
  \begin{tabular}{lll}
    \toprule
    \textit{Family} & \textit{Description} & \textit{Dim.} \\
    \midrule
    Token uncertainty & NLL mean/max, rank, $>$10 fraction, entropy slope & 4 \\
    Attention asymmetry & Self‐ratio and top-$k$ stats for three layers & $3\times3$ \\
    Residual alignment & Cosine sim. to top-attn and whole-transcript residuals & 2 \\
    Embedding similarity & Top-$k$ statistics and global max/mean & 6 \\
    Logit eigenspectrum & Log-sum singular values, spectral gap & 2 \\
    Entity grounding & Coverage ratio, novel‐entity count & 2 \\
    Surface heuristics & Trigram novelty, numeric ratio, sentence-length $z$ & 3 \\
    Semantic-graph variance & Variance of $k$-NN intra-chunk sims & 1 \\
    \midrule
    \textbf{Total} & & \textbf{35} \\
    \bottomrule
  \end{tabular}
\end{table}

Each note chunk is thus represented by a 35-dimensional feature vector $\phi(c^{Y}_{i})\in\mathbb{R}^{35}$.  We stream the vectors to a JSON-Lines file during extraction; the process is resumable, allowing the dataset to be generated incrementally on a single H200 GPU.

\begin{figure}
    \centering
    \includegraphics[width=0.98\linewidth]{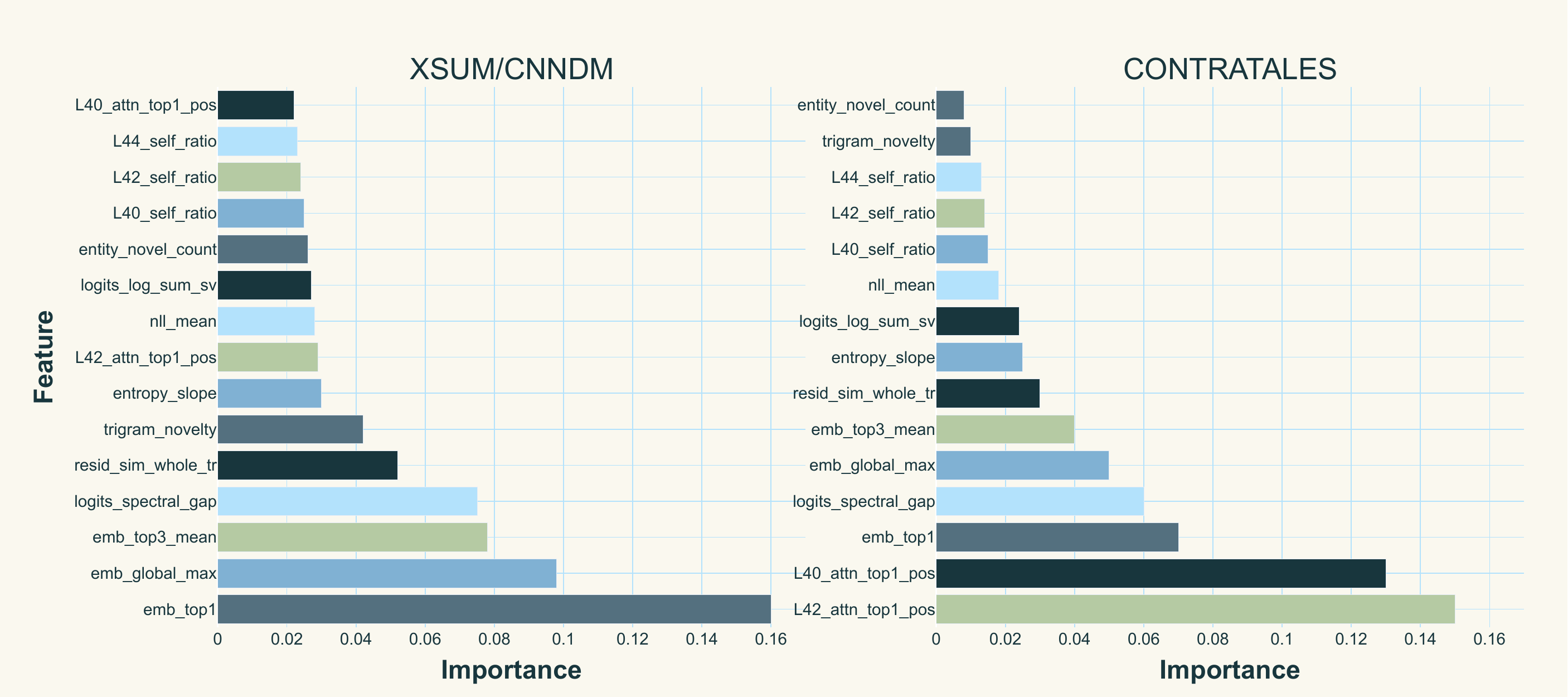}
    \caption{Comparison of top feature importances for hallucination detection across datasets. \textbf{Left}: Feature importances for detecting factual hallucinations in news summarisation datasets (\textsc{xsum}/\textsc{cnn/dm} average). Features are ranked by their importance in this scenario. \textbf{Right}: Importances of the \emph{same set of features} when applied to detecting logical contradictions in the \textsc{ContraTales} dataset. Note the significant re-ranking and change in relative importance values, highlighting how different feature types (e.g., attention-based vs. novelty-based) contribute variably to identifying distinct forms of contextual inconsistency.}
    \label{fig:feature_importance_comparison}
\end{figure}

\paragraph{Feature Importance Analysis}
To understand which signals are most indicative of hallucinations, we calculated feature importances using mean decrease in impurity (MDI) from a Random Forest classifier trained on the 35-dimensional feature vectors. Figure~\ref{fig:feature_importance_comparison} visualises the top 15 features for detecting hallucinations in the news summarisation datasets (\textsc{xsum} and \textsc{cnn/dm}, averaged) versus the \textsc{ContraTales} logical contradiction dataset.

Across both news and logical contradiction datasets, embedding-based similarity features, particularly the maximum similarity to any source chunk (\texttt{emb\_top1}) and the global maximum similarity (\texttt{emb\_global\_max}), emerge as highly predictive. This suggests that even subtle hallucinations often exhibit a detectable semantic divergence from the source material when measured by robust embedding models.

However, the relative importance of features shifts notably between the two types of datasets. For news summarisation (Figure~\ref{fig:feature_importance_comparison}, left), features like the mean of the top-3 embedding similarities (\texttt{emb\_top3\_mean}) and the spectral gap of logits (\texttt{logits\_spectral\_gap}) also rank highly. These indicate that broader semantic disconnects and unusual logit distributions are characteristic of factual hallucinations in news summaries.

In the \textsc{ContraTales} dataset (Figure~\ref{fig:feature_importance_comparison}, right), which focuses on logical contradictions, attention-based features gain considerable prominence. Specifically, the mean attention from the current note chunk to the source chunk with the highest attention in layer 42 (\texttt{L42\_attn\_top1\_pos}) and a similar feature for layer 40 (\texttt{L40\_attn\_top1\_pos}) become top-tier predictors. This highlights that logical contradictions are often signalled by how the model attends to specific, relevant parts of the source text when generating the contradictory statement. While embedding similarities remain important, their dominance is reduced compared to the news datasets.

Conversely, features like trigram novelty (\texttt{trigram\_novelty}) and the count of novel entities (\texttt{entity\_novel\_count}) show a marked decrease in importance for \textsc{ContraTales}. This is intuitive: a logical contradiction often reuses existing entities and trigrams from the source to construct the conflicting statement, making novelty a less reliable indicator than for more overt factual fabrications common in news hallucinations. The logit spectral gap also remains a strong feature for contradictions.

\section{Probe Minimal Activating Examples}
\label{app:pile}

In Table \ref{tab:pile}, we show the minimal activating examples from our hallucination probe on the Pile \citep{gao2020pile}. Most contain repetitions, either exact repetitions or semantic ones.

\begin{table}
\scriptsize
\caption{Lowest activating examples on the Pile from hallucination detection probe, trained on Gemma-2-9B. We used 1 million examples of sequence length 256 tokens from the Pile.}
\label{tab:pile}
\begin{tabular}{>{\raggedright\arraybackslash}p{0.05\textwidth}>{\raggedright\arraybackslash}p{0.40\textwidth}>{\raggedright\arraybackslash}p{0.40\textwidth}>{\centering\arraybackslash}p{0.1\textwidth}}
\toprule
\textbf{Ex \#} & \textbf{First Instance} & \textbf{Repeated Instance} & \textbf{Activation} \\
\midrule
\rowcolor{light_blue} 1 & ``Does anyone have suggestions for healy spells at low level? Apart from respeccing resto for levelling that is :)'' & ``Does anyone have suggestions for healy spells at low level? Apart from respeccing resto for levelling that is :)'' & $-30.6562$ \\

\rowcolor{light_blue} 2 & ``For a year = continued use for a year starting from the initial prescription with a gap no greater than 30 days.'' & ``For a year = continued use for a year starting from the initial prescription with a gap no greater than 30 days.'' & $-29.8125$ \\

\rowcolor{light_blue} 3 & ``Is it actually okay to treat nested std::arrays as a single flat C-style array by using .data()->data()?'' & ``Is it actually okay to treat nested std::arrays as a single flat C-style array by using .data()->data()?'' & $-28.4062$ \\

\rowcolor{lighter_blue} 4 & ``Is it possible to fill the Combobox like this programmatically?'' & ``Is it possible to fill the Combobox like this programmatically?'' & $-26.8281$ \\

\rowcolor{lighter_blue} 5 & ``Or, people in severe need of pain relief has the same need over a long time?'' & ``Or, people in severe need of pain relief has the same need over a long time?'' & $-26.8438$ \\

\rowcolor{lighter_blue} 6 & ``A pseudonym of Jehovah's end-time servant, who personifies the light that dawns on Jehovah's people at the time Jehovah restores them...'' & ``A pseudonym of Jehovah's end-time servant, who personifies the light that dawns on Jehovah's people at the time Jehovah restores them...'' & $-26.8125$ \\

\rowcolor{lighter_blue} 7 & ``[8:28 PM] cloppyhooves: They giggle, and you waifu tells you 'Oh, you won't *release* the answer? *Come* on, tell'' & ``[8:28 PM] cloppyhooves: They giggle, and you waifu tells you 'Oh, you won't *release* the answer? *Come* on, tell'' & $-26.5938$ \\

\rowcolor{lighter_blue} 8 & ``I just see how we are too far to either the left side or the right side. If we do not get back to the middle, then we will end up like Greece. The USA does not have a 'Germany' to bail them out...'' & ``I just see how we are too far to either the left side or the right side. If we do not get back to the middle, then we will end up like Greece. The USA does not have a 'Germany' to bail them out...'' & $-26.1875$ \\

\rowcolor{lighter_blue} 9 & ``Cardiff, Pembrokeshire \& South Wales'' tourism information with repeated formatting and structure & ``Cardiff, Pembrokeshire \& South Wales'' tourism information with repeated sections & $-26.1562$ \\

\rowcolor{lightest_blue} 10 & ``As for Romulan Ale I've given it some thought and I think I've come up with an idea. Ok, they call it 'ale' but it's blue.'' & Similar brewing discussion repeating terms about Romulan Ale recipe & $-25.7969$ \\

\rowcolor{lightest_blue} 11 & ``It turns out Carney was being polite when he said the caution by Canadian CEOs might be excessive. It turns out that they are in fact scaredy cats. Chickens. Nervous Nellies. Cowards, even.'' & ``It turns out Carney was being polite when he said the caution by Canadian CEOs might be excessive. It turns out that they are in fact scaredy cats. Chickens. Nervous Nellies. Cowards, even.'' & $-25.7344$ \\

\rowcolor{lightest_blue} 12 & ``It amazes me to see stuff on Amazon where the album on MP3 costs £7.99 and the cd can be bought 'new and used from' £2 or something ridiculous'' & ``It amazes me to see stuff on Amazon where the album on MP3 costs £7.99 and the cd can be bought 'new and used from' £2 or something ridiculous'' & $-25.6094$ \\

\rowcolor{lightest_blue} 13 & ``A pseudonym of Jehovah's end-time servant, whom Jehovah appoints to confront his people with their hypocrisy...'' & ``A pseudonym of Jehovah's end-time servant, whom Jehovah appoints to confront his people with their hypocrisy...'' & $-25.3906$ \\

\rowcolor{lightest_blue} 14 & ``The Bank of England has set up a research division looking at how it can get involved with digital currencies...'' & Similar content about Bank of England and digital currencies repeated & $-25.3750$ \\

\rowcolor{lightest_blue} 15 & ``What's needed is not just yet another O/RM tool (which are tuppence a dozen anyhow - I personally have written three) but a tool which supports database programming using only the conceptual model...'' & ``What's needed is not just yet another O/RM tool (which are tuppence a dozen anyhow - I personally have written three) but a tool which supports database programming using only the conceptual model...'' & $-25.2656$ \\
\bottomrule
\end{tabular}
\end{table}

\end{document}